# STATOR FLUX OPTIMIZATION ON DIRECT TORQUE CONTROL WITH FUZZY LOGIC

Fatih Korkmaz[1], M. Faruk Çakır[1], Yılmaz Korkmaz[2], İsmail Topaloğlu[1]

[1]Technical and Business Collage, Çankırı Karatekin University, 18200, Çankırı, Turkey
fkorkmaz@karatekin.edu.tr, mcakir@karatekin.edu.tr
,itopaloglu@karatekin.edu.tr

[2]Department of Electric and Electronic Engineering, Gazi University, Ankara, Turkey
ykorkmaz@gazi.edu.tr

***ABSTRACT***

*The Direct Torque Control (DTC) is well known as an effective control technique for high performance drives in a wide variety of industrial applications and conventional DTC technique uses two constant reference value: torque and stator flux. In this paper, fuzzy logic based stator flux optimization technique for DTC drives that has been proposed. The proposed fuzzy logic based stator flux optimizer self-regulates the stator flux reference using induction motor load situation without need of any motor parameters. Simulation studies have been carried out with Matlab/Simulink® to compare the proposed system behaviors at vary load conditions. Simulation results show that the performance of the proposed DTC technique has been improved and especially at low-load conditions torque ripple are greatly reduced with respect to the conventional DTC.*

***KEYWORDS***

*Asynchronous motor control, Direct torque control, Fuzzy logic optimization, Vector control*

## 1. Introduction

Nowadays, there are two well-known methods for asynchronous motor control and they can be grouped as scalar and vector control. Scalar control is based on the steady-state motor model while vector control is based on dynamic model of motor[1].

In middle of 1980's, a new vector control scheme named as direct torque control (DTC), was introduced by Takahashi[2] for variable load and speed asynchronous motor drives. It was a good alternative to the other type of vector control which known as field oriented control (FOC) due to some well-known advantages, such as simple control structure, no need much motor parameters so independency of parameter changes, fast dynamic response. Besides these advantages, DTC scheme still had some disadvantages like high torque and current ripples, variable switching frequency behavior and implementation difficulties owing to necessity of low sampling time.

Over the years, many studies have been done to overcome these disadvantages of the DTC and continues. We can group these research's under several headings:

- Research using different switching techniques and inverter topologies [3-4]
- Research using artificial intelligence on different sections of system [5-6]
- Research using different observer models [7-8]

Fuzzy Logic (FL) is one of artificial intelligence methods and its ability to incorporate human intuition in the design process. The FL has gained great attention in the every area of electromechanical devices control because of no need mathematical models of systems unlike conventional controllers[9-10]. So we also meet with the FL controller in some DTC applications. In [11], FL is used to select voltage vectors in conventional DTC and in [12] a FL stator resistance estimator is used and it can estimate changes in stator resistance due to temperature change during operation. A FL controller is used for duty ratio control method at [13]. These FL controllers can provide good dynamic performance and robustness.

In recent publications, we see that some flux optimization methods are proposed for the DTC scheme for asynchronous motor drives and the affects of the optimization algorithm is investigated. In these publications, three flux control methods are used for optimization and we can classified according to control structure as following basically: flux control as a function of torque [14], flux control based on loss model [15], and flux control by a minimum loss search controller [16].

This paper deals with a new stator flux controller on DTC scheme. It has been developed to determine the best flux reference value for motor using FL algorithm. The proposed controller self-regulates the stator flux reference without need of any motor parameters. Simulation studies have shown that this method reduces the torque ripple of the DTC scheme.

## 2. Basics of DTC

Directly control of torque by selecting the appropriate stator voltage vector has led to the naming of this method as direct torque control. The basic idea of the DTC is to choose the best vector of the voltage which makes the flux rotate and produce the desired torque. During this rotation, the amplitude of the flux remains inside a pre-defined band[17]. All measured electrical values of motor must be converted to stationary α-β reference frame on the DTC scheme and conversation matrix as given in (1-3).

$$i_{\alpha\beta 0} = [T]\; i_{abc} \tag{1}$$

$$V_{\alpha\beta 0} = [T]\; V_{abc} \tag{2}$$

$i_{abc}$, $v_{abc}$ measured and $i_{\alpha\beta 0}$, $v_{\alpha\beta 0}$ calculated phase currents and voltages respectively. $T$ is transformation matrix as given in (3).

$$T = \frac{2}{3}\begin{bmatrix} 1 & -\frac{1}{2} & -\frac{1}{2} \\ 0 & -\frac{\sqrt{3}}{2} & \frac{\sqrt{3}}{2} \\ \frac{1}{2} & \frac{1}{2} & \frac{1}{2} \end{bmatrix} \tag{3}$$

Stator flux vector can be calculated using the measured current and voltage vectors as given in (4-6).

$$\lambda_\alpha = \int (V_\alpha - R_s i_\alpha) dt \tag{4}$$

$$\lambda_\beta = \int (V_\beta - R_s i_\beta) dt \tag{5}$$

$$\lambda = \sqrt{\lambda_\alpha^2 + \lambda_\beta^2} \tag{6}$$

Where $\lambda$ is stator flux space vector, $v_{ds}$ and $v_{qs}$ stator voltage, $i_{ds}$ and $i_{qs}$ line currents in α-β reference frame and $R_s$ stator resistance. The electromagnetic torque of an asynchronous machine is usually estimated as given in (7).

$$T_e = \frac{3}{2} p(\lambda_\alpha i_\beta - \lambda_\beta i_\alpha) \quad (7)$$

Where $p$ is the number of pole pairs. An important control parameter for DTC is stator flux vector sector. Stator flux rotate trajectory is divided six sector and calculation of stator flux vector sector as given in (8).

$$\theta_\lambda = \tan^{-1}(\frac{\lambda_\beta}{\lambda_\alpha}) \quad (8)$$

Two different hysteresis comparators generates other control parameters on DTC scheme. Flux hysteresis comparator is two level type while torque comparator is tree level type. This hysteresis comparators use flux and torque instantaneous error values as input and generates control signals as output.

Switching selector unit generates inverter switching states with use of the hysteresis comparator outputs and the stator flux vector sector.

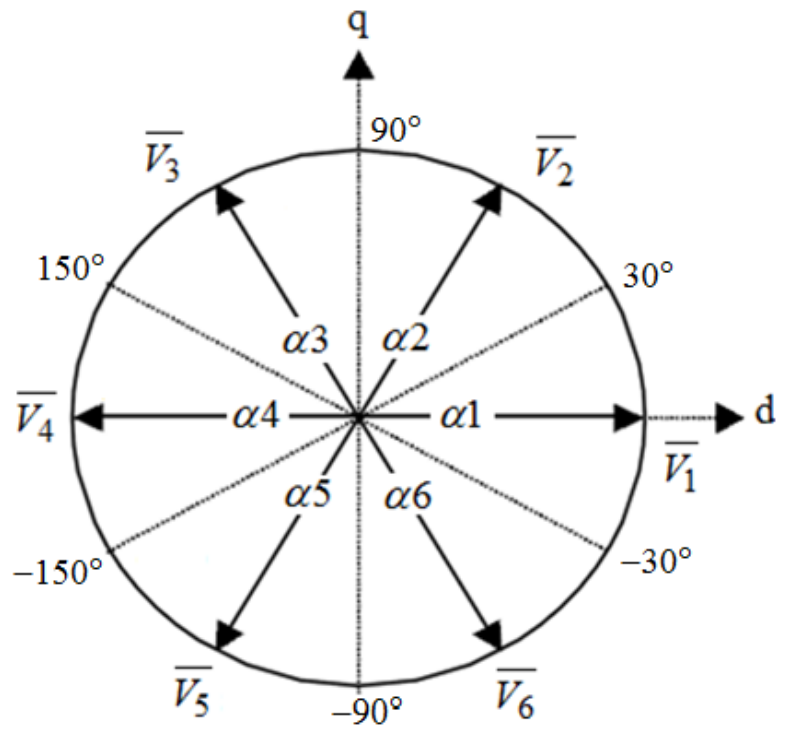

Figure 1. Inverter voltage vectors and sectors

Inverter voltage vectors and determining stator flux sector depending on the stator flux's angle are shown in Figure 1.

## 3. Fuzzy Flux Optimization Based DTC System

Conventional DTC scheme not only uses a torque reference value, but also a stator flux reference value as control parameters. Usually, motors are designed to work their maximum efficiency in their nominal operating point. But for many industrial control applications (i.e. cranes, elevators) motor loading situations can vary from time to time. Therefore, the value of motor flux should be readjusted when the load is less than the rated value. Adaptation of flux to load variations can be done in three ways: flux control as a function of torque, flux control based on loss model and flux control by a minimum loss search controller. In this paper, the first way have been preferred. It means that the flux controlled as a function of the torque but without need of any motor parameters by using fuzzy algorithm.

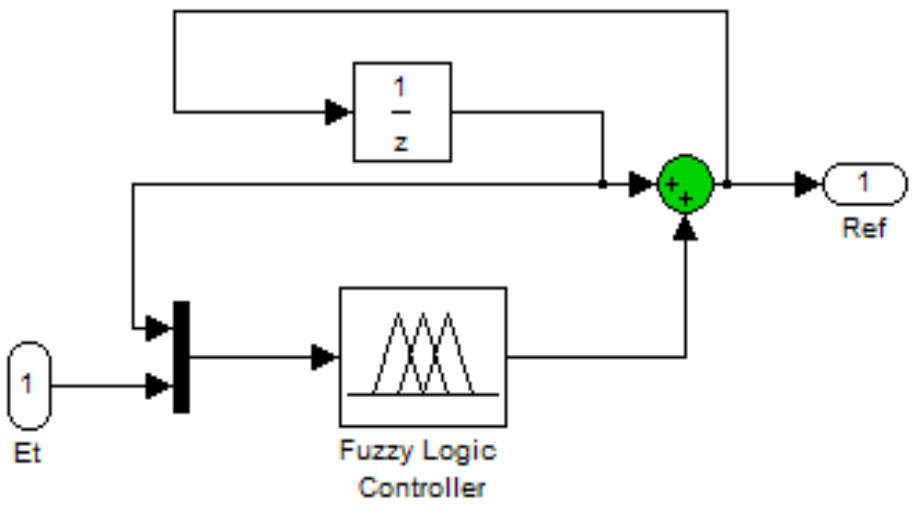

Figure 2. Fuzzy based stator flux optimization unit

The FL controller, which used in the proposed DTC scheme, utilizes the torque error and initial value of stator flux reference as control variable and generates amount of change on stator flux reference for next step as output. Fuzzy based stator flux optimization unit given in Fig 2.

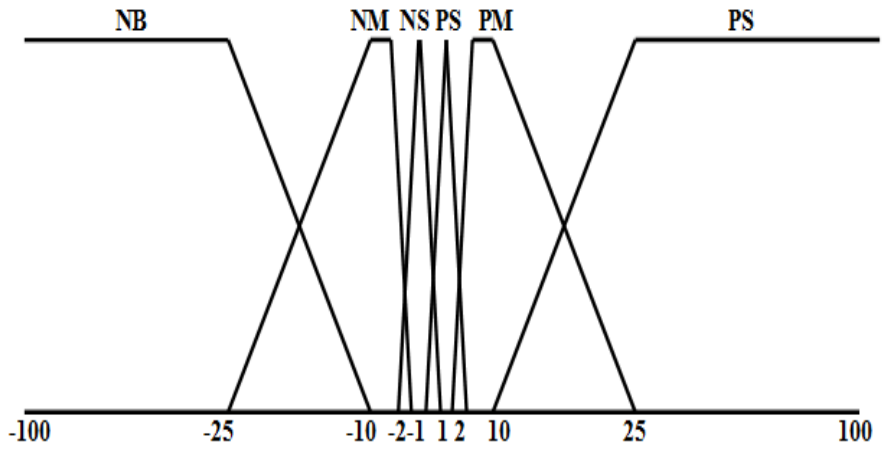

(a) Torque error

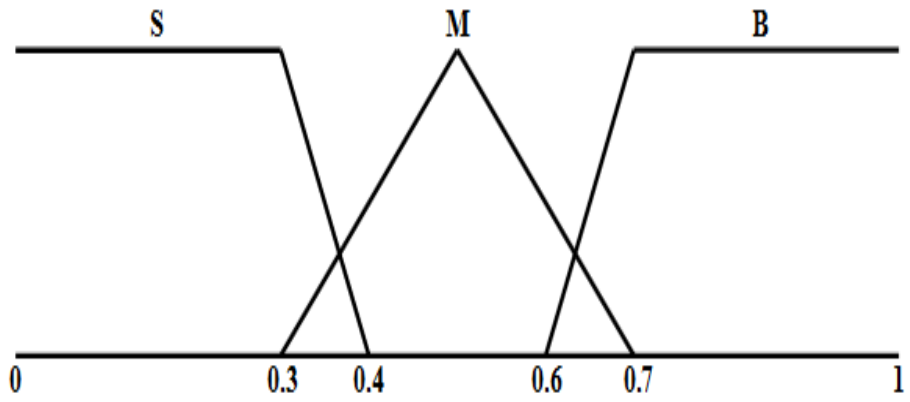

(b) Initial value of stator flux reference

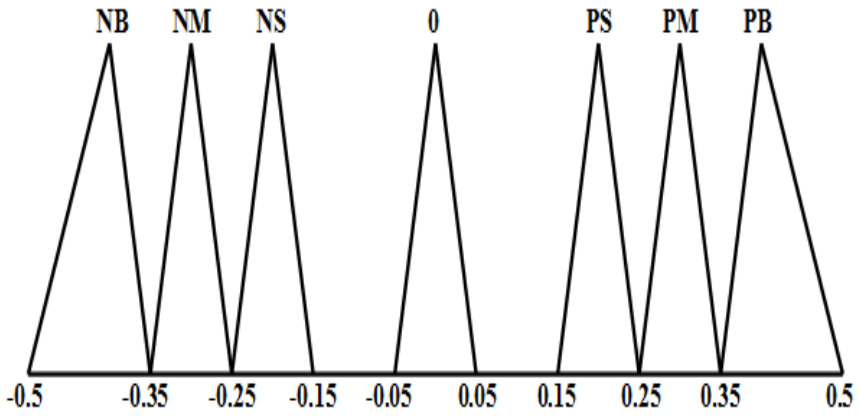

(c) Change on stator flux reference

Figure 3. Membership functions

Membership functions of the purposed fuzzy control scheme are given in Fig 3. Fuzzy control rules of the purposed fuzzy control scheme are designed to minimize torque ripples and rules can be obtained based on prior experience of investigators about the DTC scheme. FL rules shown in Table 1.

Table 1. Rule Table

| | | Torque error | | | | | |
|---|---|---|---|---|---|---|---|
| | | NB | NM | NS | PS | PM | PB |
| $\lambda_{ref}^{-1}$ | S | 0 | PS | PB | 0 | PS | PB |
| | M | NS | 0 | PM | NS | 0 | PM |
| | B | NB | NM | 0 | NB | NM | 0 |

# 4. Simulations

Numerical simulations have been carried out to investigate the effects of the proposed fuzzy stator flux controller based DTC scheme. Its developed using Matlab/Simulink®. The parameter of the asynchronous motor and simulation used in research as follows:

Table 2. Parameters of Motor and Simulations

| | |
|---|---|
| Rated Power (kW) | 4 |
| Rated Voltage (V) | 400 |
| Frequency (Hz) | 50 |
| Rated speed (rpm) | 1430 |
| Stator Resistance (Ω) | 1.405 |
| Pole pairs (*p*) | 2 |
| DC bus voltage (V) | 400 |
| Reference speed (rpm) | 1500 |
| Cycle period (µs) | 50 |

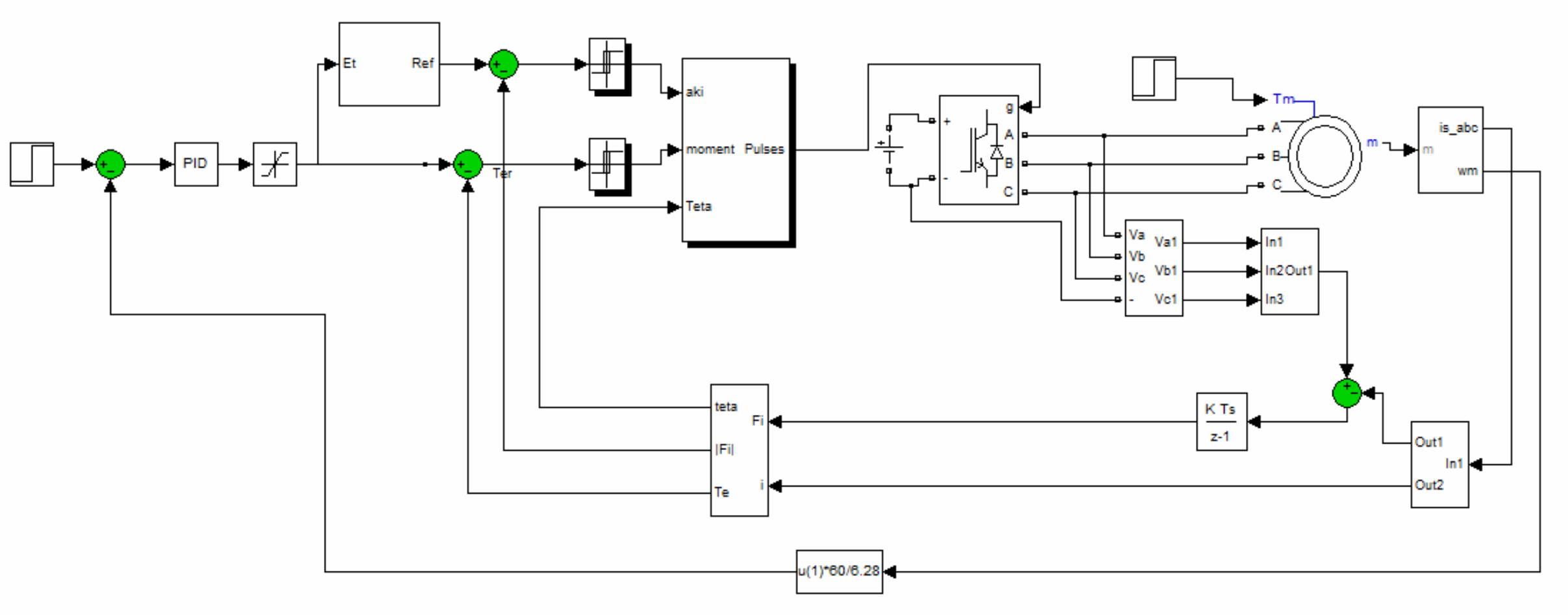

Fig.4. Simulink block diagram of the proposed DTC

At startup, the motor is unloaded, the load is changed to 10Nm at t=2 s, then load torque changed 5 Nm at t=3.5 s. for investigate the motor performance vary load conditions. The torque response curves of the conventional DTC and the proposed fuzzy stator flux optimization based DTC are shown Fig 5 and Fig 6.

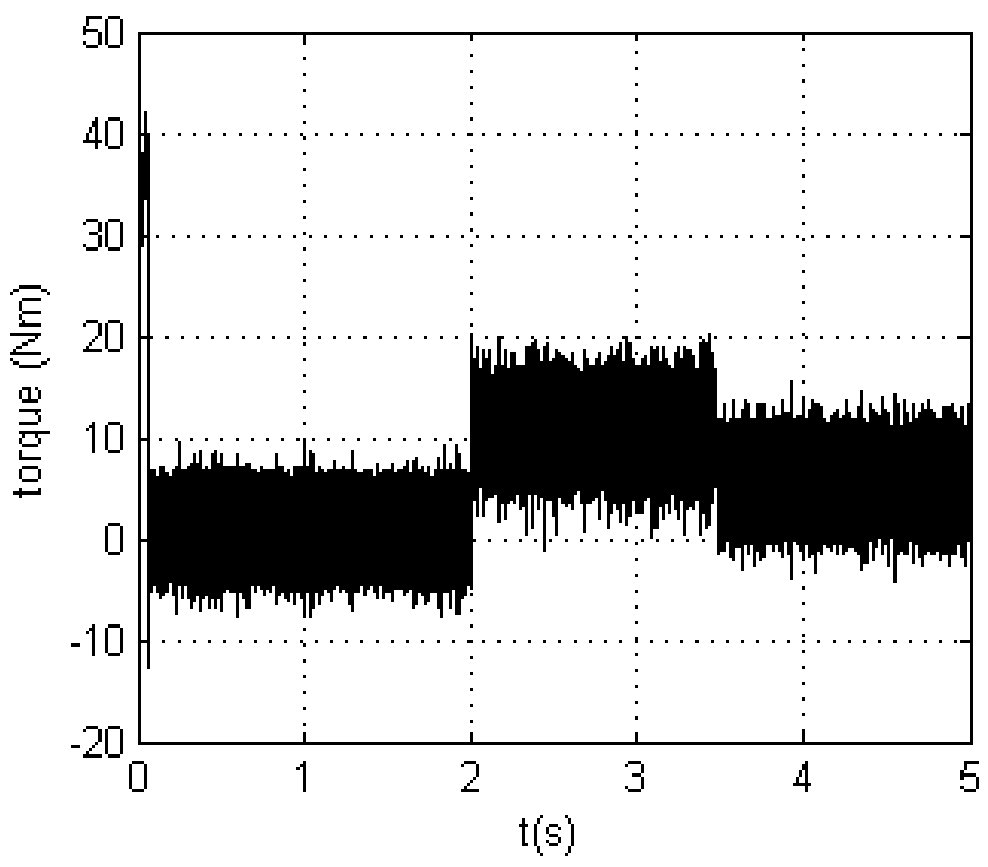

Fig. 5. Torque response of conventional DTC

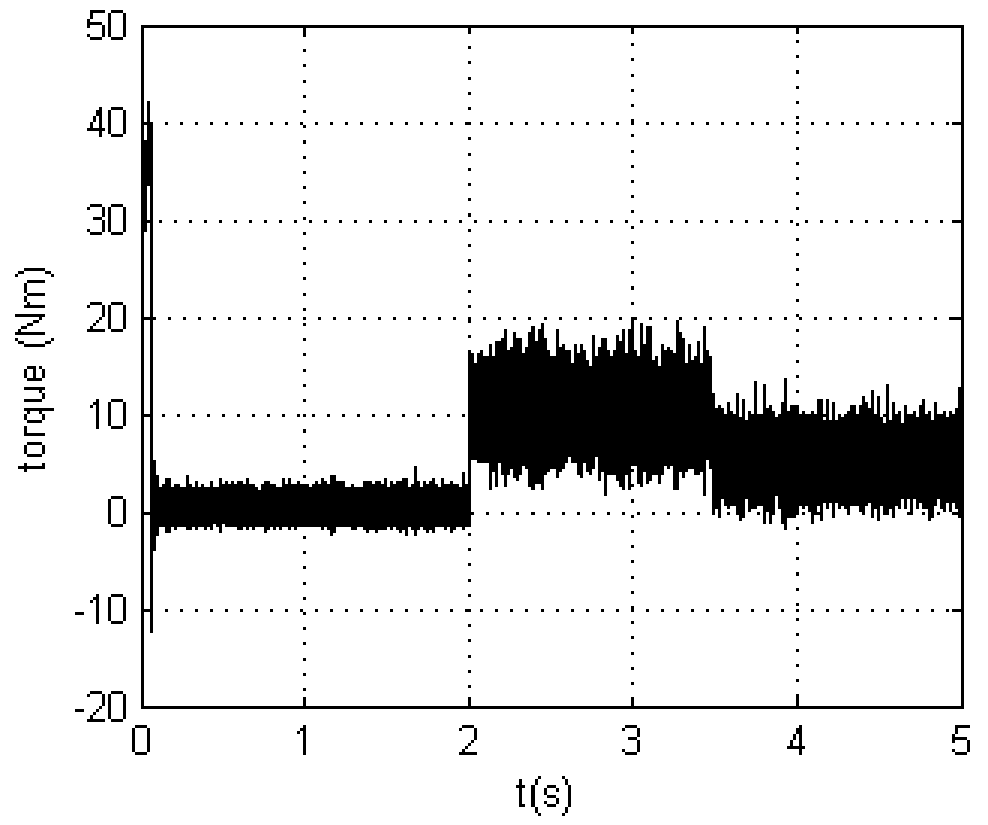

Fig. 6. Torque response of proposed DTC

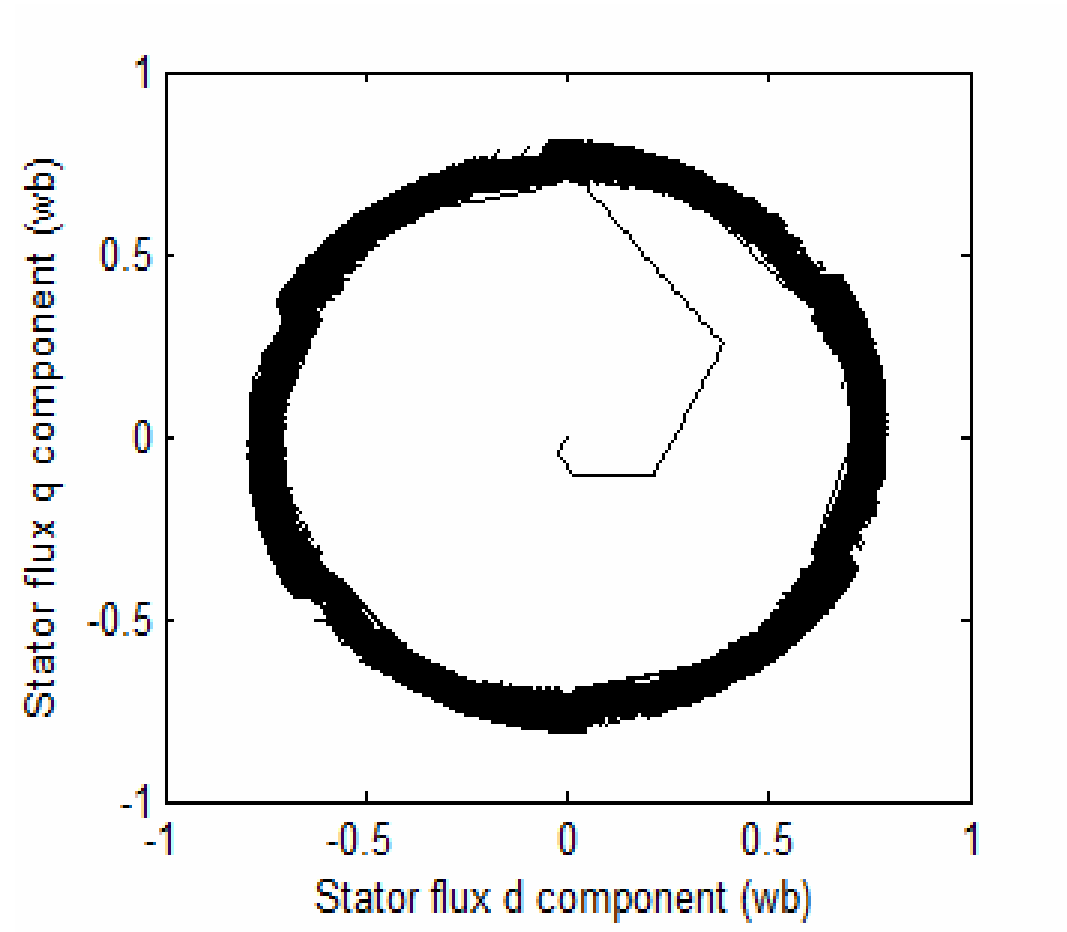

Fig. 7. Stator flux d-q components on conventional DTC

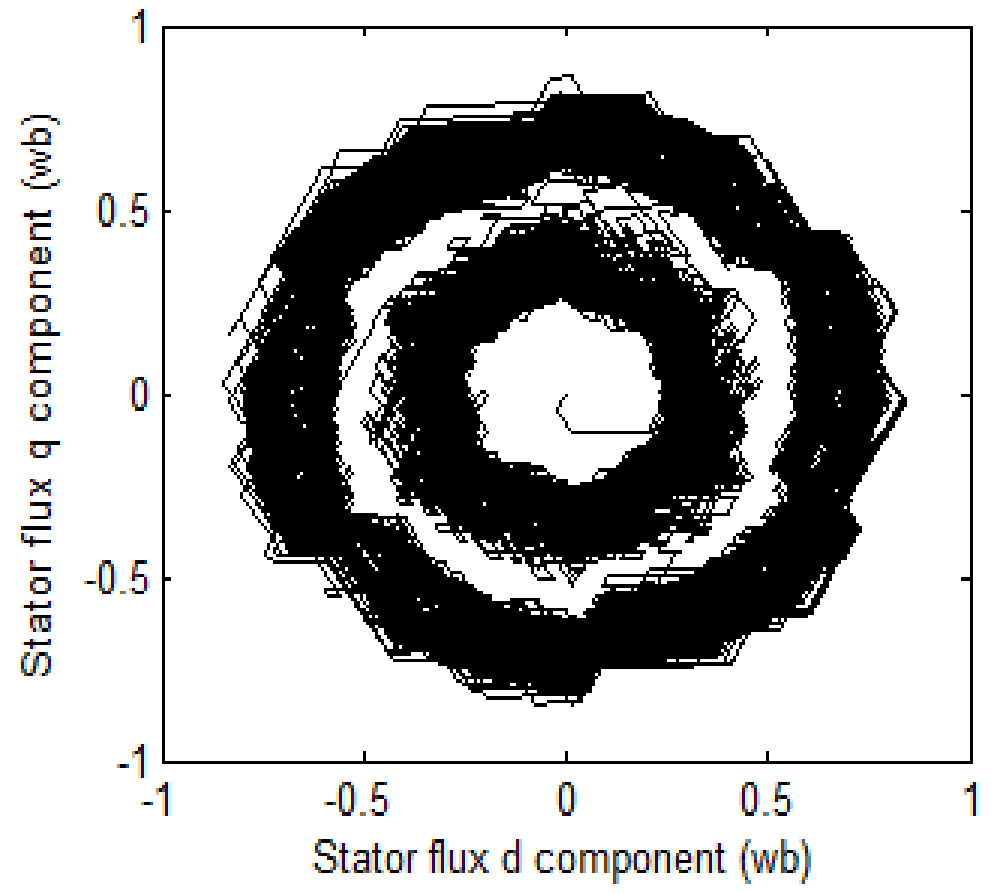

Fig. 8. Stator flux d-q components on proposed DTC

The stator flux curves of conventional DTC and the proposed fuzzy stator flux optimization based DTC are shown Fig. 7 and Fig. 8. It can be seen that the proposed stator flux optimization system finds the optimal flux value rapidly and has a better performance especially when motor load less then rated value. Obviously, the proposed system with optimized command stator flux has much smaller ripple in the torque with respect to the conventional DTC at all working conditions.

## 6. Conclusions

A new fuzzy logic based control strategy for stator flux optimization of the DTC controlled asynchronous motors has been presented in this paper. Fuzzy logic based stator flux optimizers has been designed to determine the reference value of stator flux according to torque error change without need of any motor parameter in DTC scheme. The simulation results validate that the fuzzy logic based control strategy for stator flux optimization can be successfully cooperated with conventional DTC scheme and achieves a reduction of torque ripple.

**Authors**

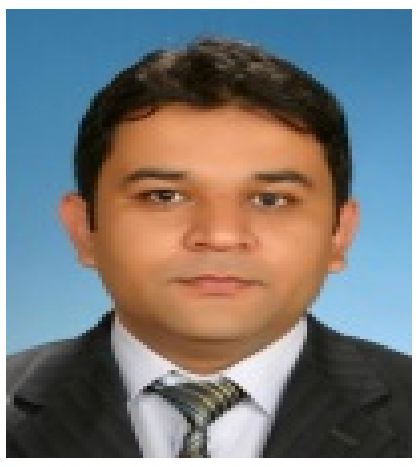

**Fatih Korkmaz** was born in Kırıkkale, Turkey in 1977. He received the B.T., M.S., and Doctorate degrees in in electrical education, from University of Gazi, Turkey, respectively in 2000, 2004 and 2011. His current research field includes Electric Machines Drives and Control Systems.

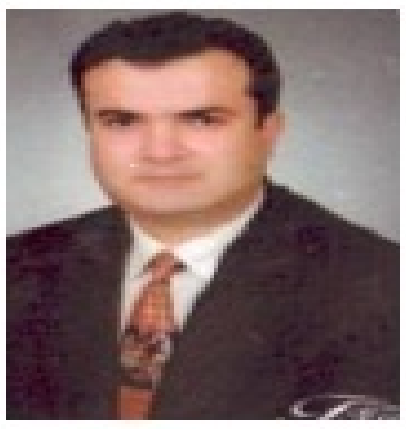

**M. Faruk Çakır** was born in Turkey in 1972. He received B.T. degree from depertmant of electric-electronic engineering, Selçuk University, Turkey, in 1994 and M.S. degree from Gebze High Technology Institute , Turkey, in 1999. Now he is PhD student in University of Gazi, Turkey. His research deals with Electric Machine Design and Nano Composite Materials.

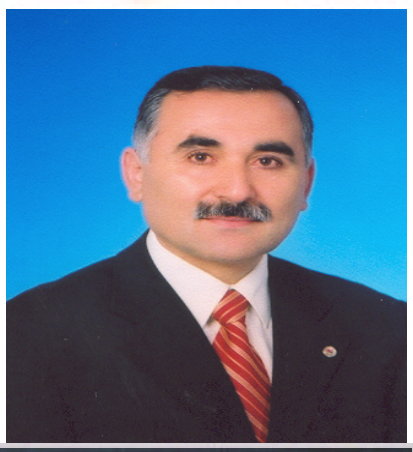

**Yılmaz Kormaz** was born in Çorum, Turkey in 1956. He received the B.T., M.S., and Doctorate degrees in electrical education from University of Gazi, Turkey, respectively in 1979, 1994 and 2005. His current research field includes Electric Machines Design and Control.

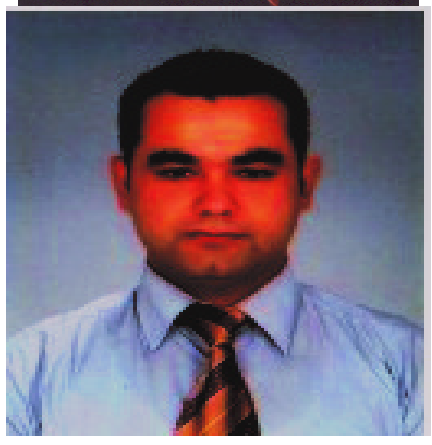

**İsmail TOPALOĞLU** was born in Adana, Turkey, in 1983. He received the B.Sc and M.Sc. degrees in electrical education from University of Gazi in 2007 and 2009, respectively. His current research interests include Computer aided design and analysis of conventional and novel electrical and magnetic circuits of electrical machines, sensors and transducers, mechatronic systems.